\documentclass[a4paper,11pt]{article}

\usepackage{float}
\usepackage{cite}
\usepackage{amsmath,amssymb,amsfonts}
\usepackage{graphicx}
\usepackage{textcomp}
\usepackage{sidecap, comment}
\usepackage{amsmath}
\usepackage{tabularx} 
\usepackage{algorithm}
\usepackage{tabulary}
\usepackage[font={small}]{caption}

\usepackage[T1]{fontenc}
\usepackage[utf8]{inputenc}
\usepackage[english]{babel}
\usepackage[font=small,labelfont=bf]{caption}

\usepackage{pgfplots}

\usepackage[english]{babel}
\usepackage[utf8]{inputenc}

\usepackage[affil-it]{authblk}
\usepackage{amsmath}
\usepackage{wrapfig}
\usepackage{xspace}
\usepackage[T1]{fontenc}
\usepackage{times}
\usepackage{fourier}
\usepackage{color}
\usepackage{url}
\usepackage[affil-it]{authblk}

\usepackage{subcaption} 

\usepackage[T1]{fontenc}
\usepackage{times}

\usepackage[noend]{algpseudocode}
\makeatletter
\def\algbackskip{\hskip-\ALG@thistlm}
\makeatother

\makeatletter
\DeclareCaptionLabelFormat{numberless}{\ALG@name#1}
\makeatother

\newlength\myindent
\begin{document}
\title{3D point cloud segmentation using GIS}
\author{Chao-Jung Liu, Vladimir Krylov \& Rozenn Dahyot}
\date{}
\affil{ADAPT Centre, School of Computer Science and Statistics,
Trinity College Dublin, Ireland}
\maketitle
\thispagestyle{empty}

\begin{abstract}
In this paper we propose an approach to perform semantic segmentation of 3D point cloud data by importing the geographic information from a 2D GIS layer (OpenStreetMap). The proposed automatic procedure identifies meaningful units such as buildings and adjusts their locations to achieve best fit between the GIS polygonal perimeters and the point cloud. Our processing pipeline is presented and illustrated by segmenting point cloud data of Trinity College Dublin (Ireland) campus constructed from optical imagery collected by a drone.         
\end{abstract}


%%%%%%%%%%%%%%%%%%%%%%
\section{Introduction}

%%%%%%%%%%%%%%%%%%%%%%
Up-to-date accurate 2D and 3D maps are growing increasingly important for localisation and navigation employed by both humans and machines. Various technologies and data collection modalities are available nowadays to capture and encode a digital twin of the world such as Lidar \cite{DublinLidar15} and drone imagery \cite{TCDDataset2017}. Once created, such digital twins can be seamlessly manipulated and visualised with the help of game engines, and used in applications, such as education (e.g., driving simulators), and entertainment (e.g.,  virtual visits and gaming \cite{10.1007/978-3-319-70111-0_29}).  
More recently these virtual environments have also found applications in providing valuable labeled data for training machines  using data driven artificial intelligence. For instance project Airsim \footnote{https://github.com/Microsoft/AirSim} uses the Unreal game engine to provide training data for  autonomous drones and cars.

\begin{figure}[!t]
  \begin{subfigure}[]{0.5\textwidth}
     \includegraphics[width=\textwidth, height=5cm]{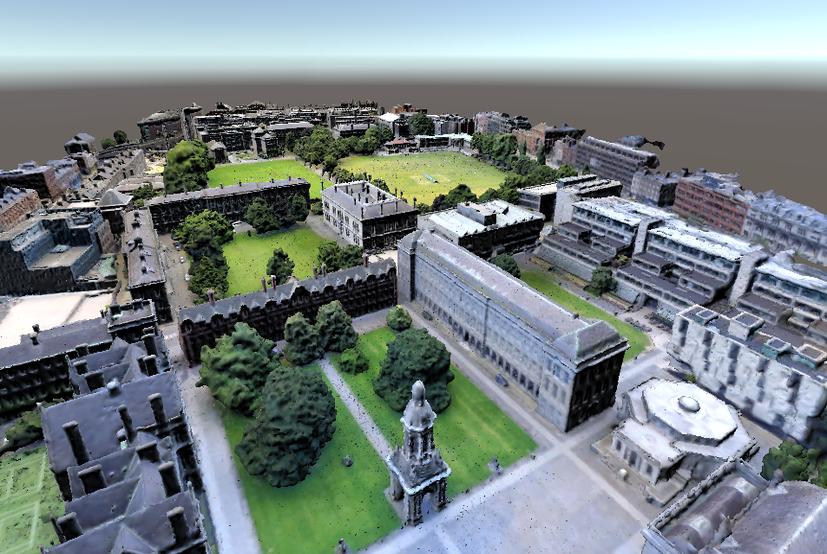}
    \caption{Trinity 3D model dataset \cite{TCDDataset2017}}
    \label{fig:TCD}
  \end{subfigure}
  \begin{subfigure}[]{0.5\textwidth}
    \includegraphics[width=\textwidth, height=5cm]{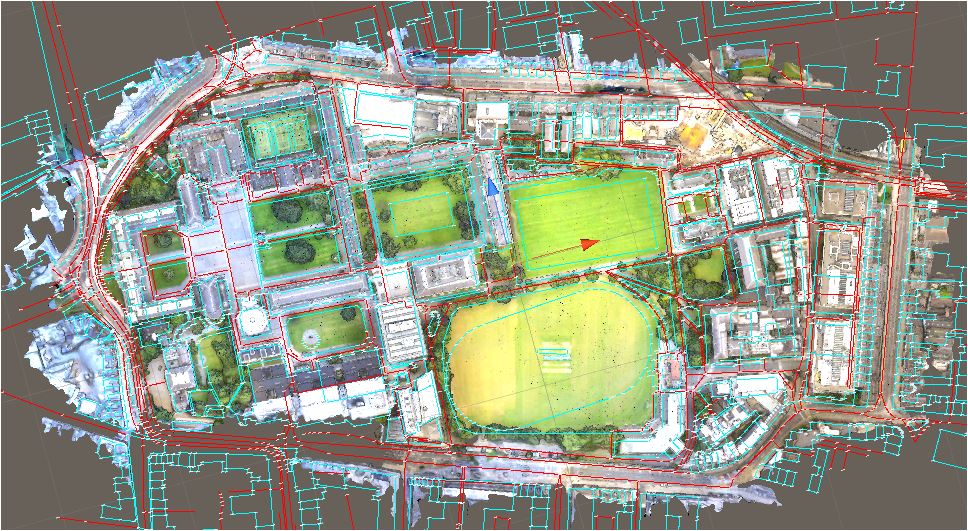}
    \caption{Overlay OSM data onto game terrain}
    \label{fig:Overlay}
  \end{subfigure}
  \caption{3D point cloud of Trinity College Dublin campus, 2015.}\label{fig:problem}
\end{figure}

In this paper, our aim is to label automatically unstructured geolocated 3D point cloud data. For this purpose we propose to register heterogeneous sources of information: the semantic  information  provided by  OpenStreetMap (OSM) and  3D point clouds covering the same geographic area (Fig. \ref{fig:problem}). We show that the registration of these two sources of information allows one to segment the point cloud in useful semantic units with arbitrary geometrical shapes. After introducing some related work (Sec. \ref{sec:SOA}), our approach  is presented in Section \ref{sec:our:method}.

%%%%%%%%%%%%%%%%%%%%%%%%%
%%%%%%%%%%%%%%%%%%%%%%%%%
\section{Related Work}
\label{sec:SOA}

Structure from motion techniques have been widely used for reconstruction of large scale scenes and image geo-location. Practically, it has become a cheaper alternative to costly high-end LiDAR technology. Structure from motion relies on keypoint detection, image matching, bundle adjustment and, finally, generation of dense point clouds. These points can then be condensed into meshes and be presented in game engine as terrain (cf. Fig. \ref{fig:TCD}).
For instance, Agarwal et al. \cite{building-rome-in-a-day} use a large collection of pictures harvested from the web to create a sparse point cloud of a city. 
 B\'{o}dis-Szomor\'{u} et al.~\cite{DBLP:journals/corr/Bodis-SzomoruRG16} proposed to reconstruct a mesh of a city area by registering and fusing two point clouds, one airborne and  the other generated from street view and aerial images to reconstruct the city view, and then generate the mesh by using fusion techniques.
Rumpler et al. \cite{de4ce0e7de2d473bbb2e4add98609ee6}  register  OSM data for retrieving the information of outlines of buildings in the form of 2D vectors, and use them with satellite pictures to generate 3D mesh building. 
Bulbul and Dahyot \cite{BULBUL201728} reconstruct cubical model of a city using OSM  geolocated building footprints  and heights, and associate  textures  extracted from Google Street view imagery to recover building facade. The resulting 3D city model is then used in a game engine for visualisation of   social media activity in the area, using  pictures posted on social media platforms. 

There is a large body of work for image region labeling into generic categories such as cars, trees or buildings. 
For instance, Tighe and Lazebnik \cite{10.1007/978-3-642-15555-0_26} proposed an  image parsing method to segment the region of images by trained descriptors. The semantic labels are classified by predefined class for geometric / semantic context.
The semantic labeling work is conducted through all of the overlapping images  for reconstruction, which is very time consuming. 
Having a 3D mesh generated multiple view imagery, 
Riemensschneider et al. \cite{10.1007/978-3-319-10602-1_34} improve computation complexity and accuracy by predicting the  best  image view for clustering and propagate the labels to the mesh. It is shown to be more efficient that than clustering of all images and merging labels for a same object in the scene. Tchapmi et al. \cite{tchapmi_segcloud_3dv17} label raw point clouds using  neural networks  with Conditional Random Field  to predict class of each point in the cloud.
Kaiser et al. \cite{Kaiser2017LearningAI} segment the 2D aerial imagery into object by using the data from OSM, transforming geographic coordinate to local pixel coordinates, and use neural network architecture to refine the boundary to make a finer-grained labeling for objects.
Becker et al.~\cite{isprs-annals-IV-1-W1-3-2017} proposed to classify photogrammetric  3D point  clouds into specific classes (e.g. road, building, vegetation). 

Krylov et al. \cite{rs10050661} proposed to geolocate street furniture from Google Street View (GSV) imagery  using fully convolutional neural networks for object segmentation and distance estimation followed by a Markov Random Field to coherently geolocate individual objects in images. This was extended in~\cite{KrylovICIP18} to the fusion of street level imagery and LiDAR data to object detection at increased spatial accuracy.
%Using GSV imagery  only allows to locate objects within the standard GPS sensor accuracy range of approximately 2 meters. 
Similarly, Branson et al.~\cite{BRANSON201813} use  GSV images together with optical satellite imagery for cataloguing street trees.

%%%%%%%%%%%%%%%%%%%%%%%%%%%%%%%%%%%%%%%%%%%%
%%%%%%%%%%%%%%%%%%%%%%%%%%%%%%%%%%%%%%%%%%%%
\section{Geolocated Data Registration}
\label{sec:our:method}

Our approach is demonstrated using the  data  for Trinity College Dublin Campus recorded using a drone in 2017, which is available open source \cite{TCDDataset2017}. Our methodology employs  the 3D point cloud generated from drone imagery (Sec. \ref{sec:drones}) but can also be applied to Lidar data (e.g \cite{DublinLidar15}).
We employ OSM data as GIS source of information \cite{de4ce0e7de2d473bbb2e4add98609ee6,Kaiser2017LearningAI,BULBUL201728}. 

\subsection{3D point clouds generated from drone imagery}
\label{sec:drones}

The Trinity campus mesh  has been  generated from optical imagery captured by a drone \cite{2017JARS...11b5015B} in 2017. Pix4D software\footnote{https://pix4d.com/} has been employed to process the images to generate the geolocated point cloud using structure from motion techniques.

\subsection{OSM data structure and parsing}

To acquire geolocation and semantic label data, we employ the OSM API that allows one to extract information about specific geographic area from their database in XML format. We use this interface to collect semantic information about the Trinity campus. The XML dump file contains all 3 types of OSM attributes: ways, nodes and relations. Ways are used to encode polygon-shaped areas like buildings, roads, etc. A way contains semantic labels (tags) and references to corresponding nodes. Each node contains the geolocation (longitude and latitude) of an individual point on the map (e.g., corner of a building). Relations are used to model local logical or geographic relationships between objects. Using the name of any individual building we can search among all the way attributes to find all the nodes establishing its location inside the Trinity campus.

\subsection{Mercator projection}
OSM uses WGS84 spatial reference system. This is the reference system also used by GPS. The corresponding coordinates are referred to as longitude and latitude. WGS84 models Earth as a spheroid. The real shape is however ellipsoidal, i.e. flat at both poles. To resolve this inconsistency, the Mercator projection\footnote{https://wiki.openstreetmap.org/wiki/Mercator} is employed.

\begin{figure}[!t]
\begin{center}
    \includegraphics[width=.5\linewidth]{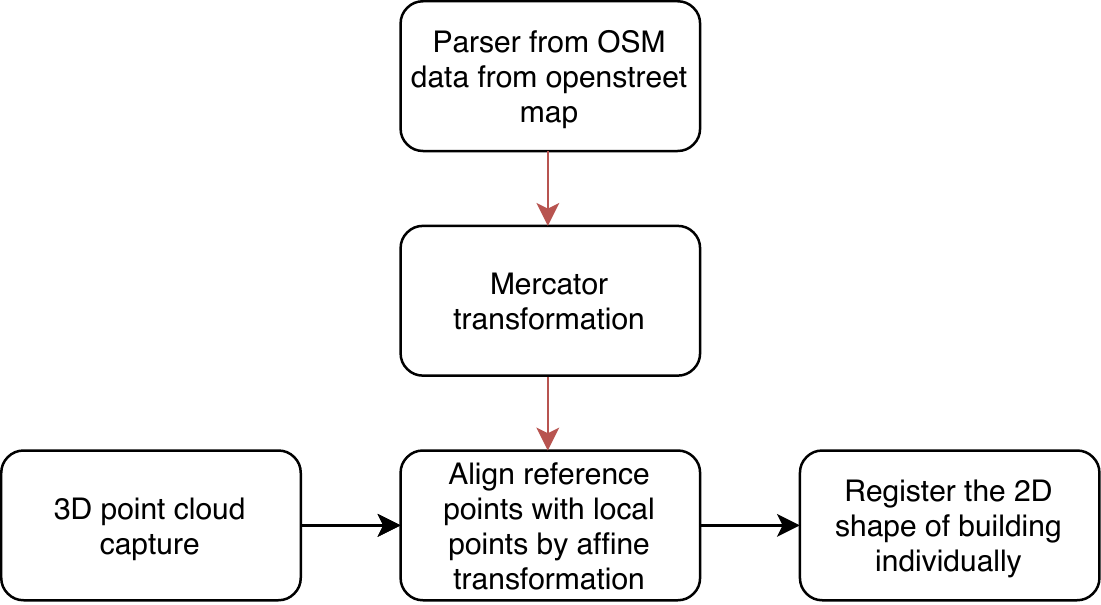}
    \caption{Registration is performed in two steps: affine registration for the whole dataset, followed by a separate position adjustment for each building. }\label{fig:pipeline}
    \end{center}  
\end{figure}

\begin{figure}[!h]
\centering
 \begin{subfigure}[]{0.58\textwidth}
\includegraphics[width =1\textwidth]{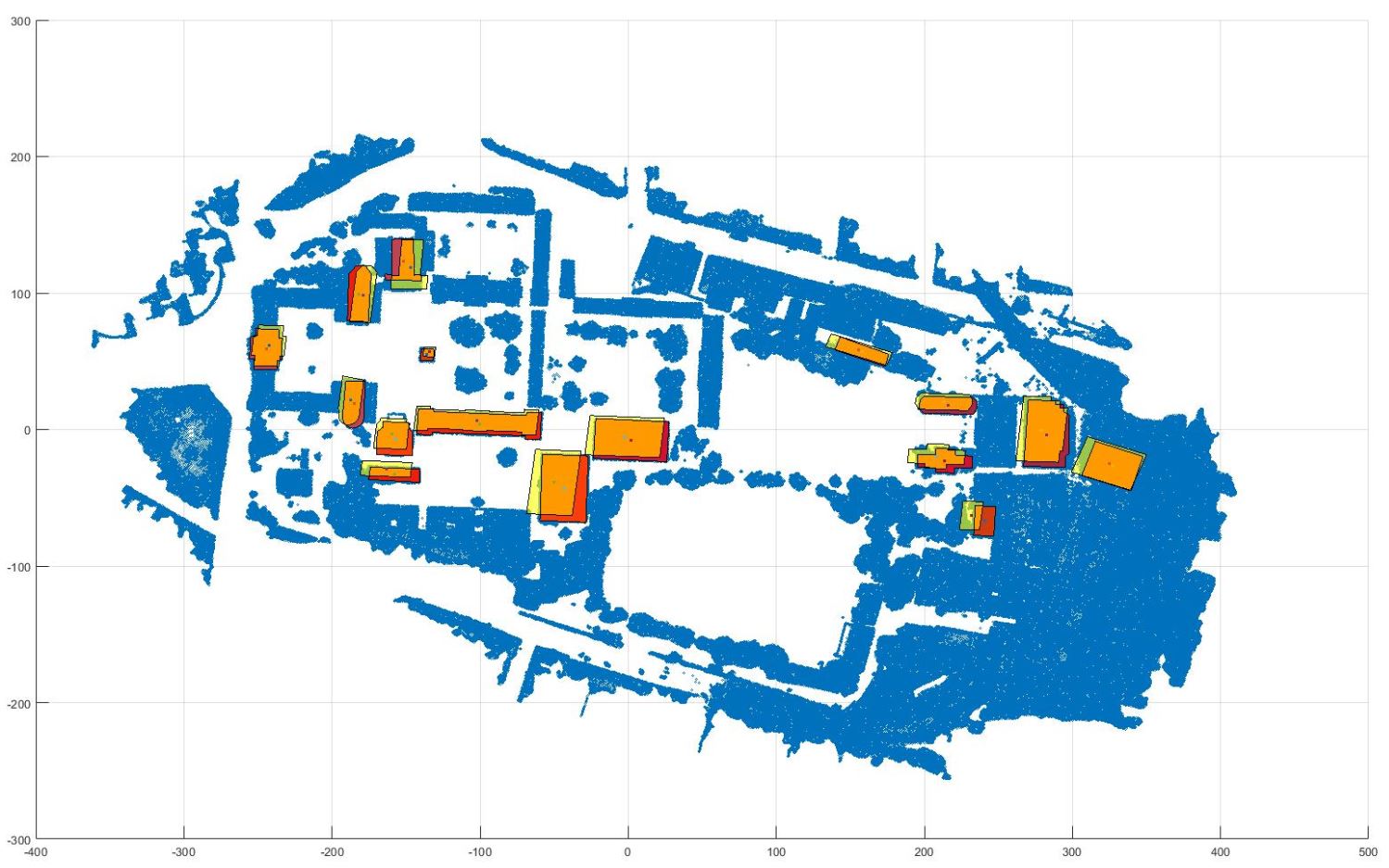}
\caption{Intersection (orange)  between OSM building positions (yellow) and 3d point cloud ground truth locations (red).}\label{fig:IOU}
\end{subfigure}\quad
\begin{subfigure}[]{0.38\textwidth}
\begin{center}
\begin{tabular}{ccc}
    \hline
    \hline       
      & IoU  & Distance (meters) \\ 
    \hline           
    \hline
    Average &0.567&5.594  \\
    Median &0.607&4.871 \\ 
    \hline 
    \end{tabular}
    \end{center}
      \caption{Distance between centroids defined by OSM and 3d point cloud across 16 buildings.} 
      \label{Table:firstIoU}
      \end{subfigure}%
      \caption{Fit between point cloud data and OSM GIS data after global registration. }
\end{figure}

\subsection{User defined correspondences between heterogeneous data streams}

To align the reference location from Mercator projection to 3D model, we manually selected  18  correspondences (control points) between the two media streams 
to estimate an affine transformation matrix  handling translation, rotation and scale with Least Squares.

The center of Trinity campus model (0,0)is on the Museum building, has  OSM coordinate (53.34380,-6.25532) and Mercator projection coordinate(-696339.0371489801,7012543.77625507). 

Fig. \ref{fig:pipeline} presents our processing pipeline. 

\subsection{Evaluation of Global Affine registration with user defined correspondences}

To evaluate how accurate the geo-locations of nodes from OSM fit the considered point cloud dataset, we report assessment on 16 buildings across the campus. We also manually label the point cloud by photo-interpretation to have a ground truth of semantic units corresponding to these 16 buildings. 
Each building is defined by its perimeter, i.e. a polygon shape on x-y plane in orthographic view (see red polygons in Fig. \ref{fig:IOU}). 
The information about building locations coming from OSM is reported as yellow polygons in Fig.~\ref{fig:IOU}. We compare theses two polygons as 2D shapes (disacrding the third dimension of data) to assess the correspondance via overlap using {\it Intersection of Unions} (IoU)~\cite{gabriela2013WhatIA}.
The IoU is a real number between 0 to 1, where higher values mean better fit (equivalently, stronger overlap). 
We compute the shape's centroid points and use these to find distance between polygons. 

\begin{figure}[!b]
\begin{tikzpicture}
    \begin{axis}
        [
        ymin=0, ymax=1,
        ,width=\textwidth
        ,height = 4cm
        ,ylabel=IoU
        ,xtick=data,
         xticklabel style = {rotate=45, anchor=east, align=right,text width=1.5cm,font=\tiny,yshift=-.5ex},
        ,xticklabels={Chapel, Regent House, Berkeley Library, O'Reilly, Museum Building, Botany Building, Campanile, Trinity Long Room Hub, The Old Dining Hall, Reading Room, Old Civil Eng. Building, FitzGerald Building, Lloyd Institute, Zoology Building, Old Library, Public Theather}
        ]
      \addplot+[
        only marks,
        scatter,
        mark=*,
        mark size=2.5pt]
    	coordinates
		{(0,0.6231)(1,0.7278)(2,0.5433)(3,0.7633)(4,0.7163)(5,0.5837)(6,0.436)(7,0.3455)(8,0.3909)(9,0.5994)(10,0.6906)(11,0.4867)(12,0.6919)(13,0.1562)(14,0.705)(15,0.6155)};
    \end{axis}
    
\end{tikzpicture}
\caption{IoU per building after global affine registration.}\label{plot:firstIoU}
\end{figure}
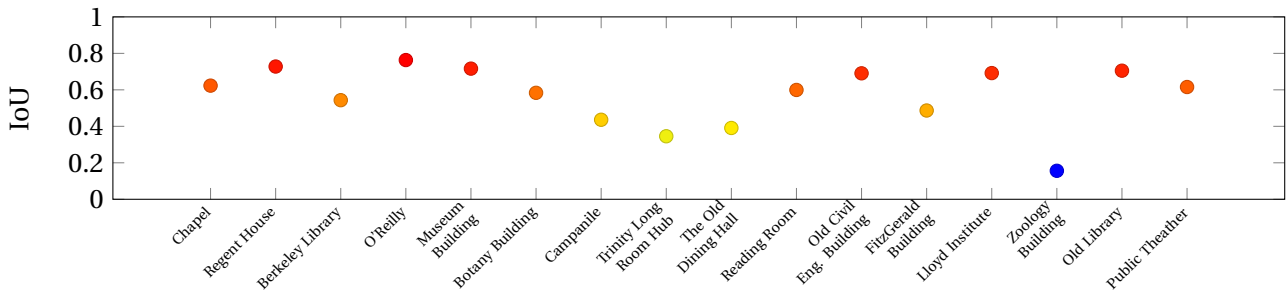

As can be seen (Fig. \ref{fig:IOU}), there is a substantial misalignment between geolocations reported by OSM and 3d point cloud. 
This is mainly because one single affine transformation does not capture well the deformation between the two streams of information. Furthermore, the location information in OSM is crowd-sourced and is not always of the highest spatial accuracy.
Figure~\ref{Table:firstIoU} reports the averages and median values for IoU and the distance between OSM and 3d point cloud polygons' centroids.
%The ratio of this Trinity model is 9.38  to 10.83 meters in the real world. 
%TABLE \ref{Table:firstIoU} describes how well GIS data register with priors. 
Fig. \ref{plot:firstIoU} demonstrates the IoU for each of building and it reflects to the misalignment for each of building.

% We plot out few more examples for each of building.
% Fig. \ref{raw} and Fig. \ref{raw3} would be considered as poor registration, while Fig. \ref{raw2} and Fig. \ref{raw4} both are good registration. However, the overall result is not good enough to apply to any application. The average error is equivalent to 5.6 meters in real world.
% \begin{minipage}{.5\linewidth}
% \begin{figure}[H]
%   \begin{subfigure}[]{0.38\textwidth}
%     \includegraphics[width=\textwidth, height=4cm]{pics/raw.JPG}
%     \caption{Zoology Building IoU:0.1562}
%     \label{raw}
%   \end{subfigure}
%   %
%   \begin{subfigure}[]{0.38\textwidth}
%     \includegraphics[width=\textwidth, height=4cm]{pics/raw2.JPG}
%     \caption{Regent House IoU:0.7278}
%     \label{raw2}
%   \end{subfigure}
%   %
% \end{figure} 
% \begin{figure}[H]  
%    \begin{subfigure}[]{0.38\textwidth}
%     \includegraphics[width=\textwidth, height=4cm]{pics/raw3.JPG}
%     \caption{Trinity Long Room IoU:0.3455}
%     \label{raw3}
%   \end{subfigure}
%     \begin{subfigure}[]{0.38\textwidth}
%     \includegraphics[width=\textwidth, height=4cm]{pics/raw4.JPG}
%     \caption{Chapel IoU:   0.6231}
%     \label{raw4}
%   \end{subfigure}
%   \caption{}
% \end{figure}

% \par
% \end{minipage}

%%%%%%%%%%%%%%%%%%%%%%%%%%%%%%%%%%%%%%%%%%%%
\section{Adjustment of OSM locations}

Alongside the considered 16 buildings our 3d point cloud is comprised of other points, describing ground, trees and other buildings. We first remove the ground points by elevation surface thresholding using CloudCompare(available at \url{https://www.danielgm.net/cc/}). We then discard the third coordinate data in the 3d point cloud (elevation above the reference level) transforming it into 2d point cloud $D_{2d}$.

We employ IoU as the main metric to assess the quality of fit between the data and ground truth locations.
In order to improve IoU for buildings, we propose to estimate the optimal translation from OSM-defined positions towards 3d point cloud clusters. We assume that the spatial orientation (rotation) and scaling are correct in OSM.
In other words, to identify parts of the point cloud that correspond to specific buildings we assume that the polygon defining the perimeter of the building is recovered by affine translation of the OSM position within a certain maximum radius.

\begin{comment}
\begin{algorithm}
\caption{Adjust OSM position}
\begin{algorithmic}[1]
\Procedure{function $f$}{}
\BState{$hypervariable$}:
\State{Floating gap for every looping time} : $\Delta_i$
\State{Looping times} : $K_i$
\State{Threshold distance} : $T_d$
\State{Threshold IoU} : $T_I$
\BState{For each building }:
\BState{$Init$}:
\State{Select region of interest($r$)};
\State{Retrieve 2d vector $S_i$ and compute centroid of $C{S_i}$};
\State{Initialize number of points clustered: $P$};
\BState \emph{loop}:
\While{not the end of $K_i$ }
\State{Searching from top-left, floating by $\Delta_i$};
\State{$C{T_{x,y} S_i}$ $\leftarrow$ Compute centroid of $T_{x,y} S_i$ };
\If{$\mid$ $C{S_i}  - C{T_{x,y} S_i}$ $\mid$ $<$ $T_d$}
\State{$P'$ $\leftarrow$ Compute number of points clustered}
\ElsIf{$P'$ $>$  $P$ AND $IoU$ > $T_I$ }
\State $P\leftarrow P'$
\Else
\State{\emph{goto loop}}
\EndIf
\EndWhile

\EndProcedure{End procedure}
\end{algorithmic}
\end{algorithm}
\end{comment}

\begin{figure}[!t]
  \begin{subfigure}[]{0.33\textwidth}
    \includegraphics[width=\textwidth, height=5cm]{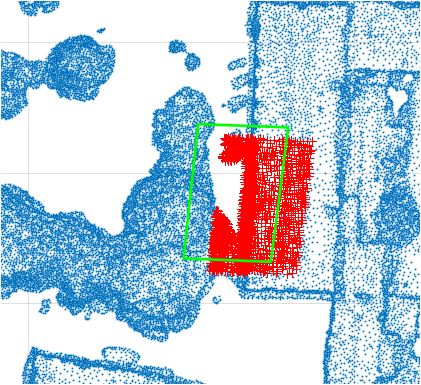}
    \caption{Zoology bld., $r_3=5.5m$}
    \label{fig:Process_Zoology_a}
  \end{subfigure}
  \begin{subfigure}[]{0.33\textwidth}
    \includegraphics[width=\textwidth, height=5cm]{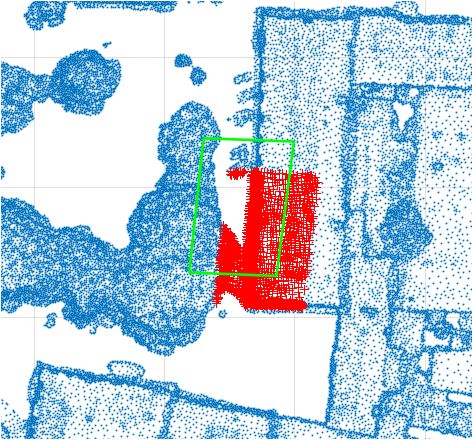}
    \caption{Zoology bld., $r_2=8m$}
    \label{fig:Process_Zoology_b}
  \end{subfigure}
   \begin{subfigure}[]{0.33\textwidth}
    \includegraphics[width=\textwidth, height=5cm]{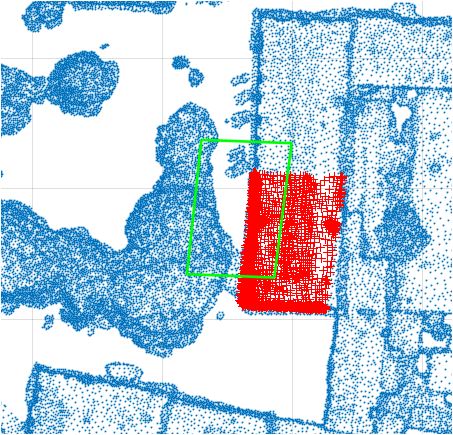}
    \caption{Zoology bld., $r_1=12m$}
    \label{fig:Process_Zoology_c}
  \end{subfigure}\\
  \begin{subfigure}[]{0.33\textwidth}
    \includegraphics[width=\textwidth, height=3.8cm]{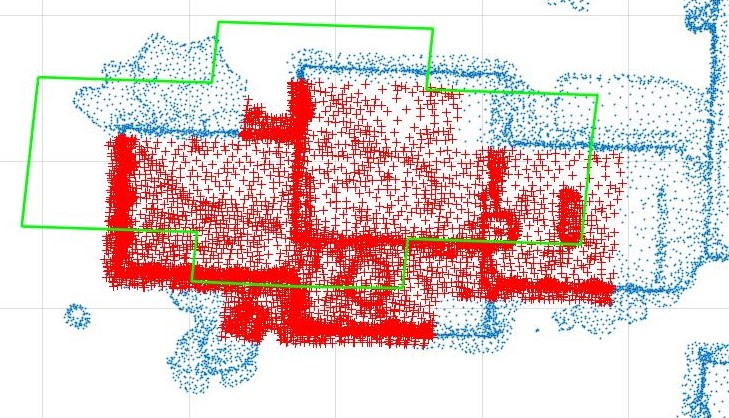}
    \caption{FitzGerald bld., $r_3=5.5m$}
    \label{fig:Process_FitzGerald_a}
  \end{subfigure}
  \begin{subfigure}[]{0.33\textwidth}
    \includegraphics[width=\textwidth, height=3.8cm]{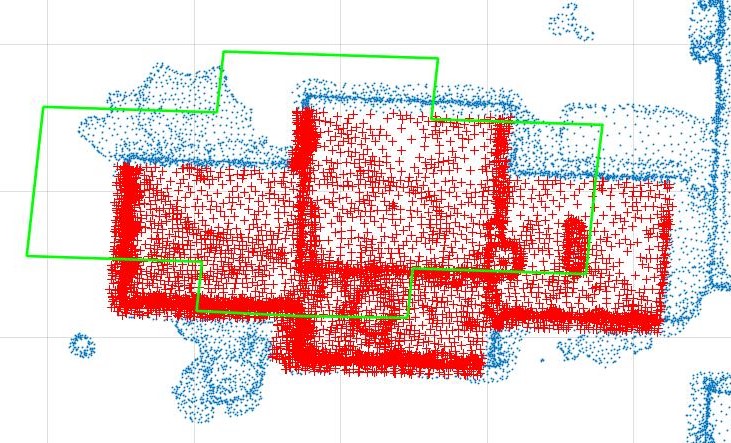}
    \caption{FitzGerald bld., $r_2=8m$}
    \label{fig:Process_FitzGerald_b}
  \end{subfigure}
   \begin{subfigure}[]{0.33\textwidth}
    \includegraphics[width=\textwidth, height=3.8cm]{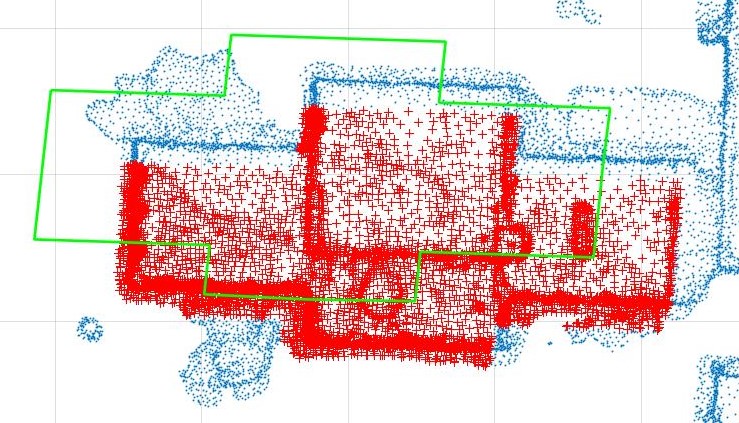}
    \caption{FitzGerald bld., $r_1=12m$}
    \label{fig:Process_FitzGerald_c}
  \end{subfigure}
    \caption{2D point cloud after ground removal (blue points) and the segmentation (red points) obtained by the proposed OSM adjustment procedure for various search radii $r$ exemplified on Zoology and FitzGerald buildings. Green polygons represent the original OSM building positions.}
    \label{buildings}
\end{figure}

\begin{table}[!b]
\centering
\begin{tabular}{l|cc|cc|cc}
& \multicolumn{2}{c|}{$r_1=12m$} & \multicolumn{2}{c|}{$r_2=8m$} & \multicolumn{2}{c}{$r_3=5.5m$} \\
Metric    & IoU & Distance & IoU & Distance & IoU & Distance \\
\hline
Average ($\Delta_1=0.8m$) & 0.791    & 2.411  & 0.774   & 2.269  & 0.764   & 2.577  \\
Median ($\Delta_1=0.8m$)  & 0.779    & 2.232  & 0.801   & 2.128  & 0.773   & 2.235 \\
Average ($\Delta_2=1.15m$) & 0.791    &  2.266  & 0.784   & 2.197  & 0.759   & 2.806  \\
Median ($\Delta_2=1.15m$)  & 0.787    & 2.065  & 0.799   & 2.056  & 0.783   & 2.876 
\end{tabular}
\caption{Performance of the proposed position adjustment method for different values of $r$: IoU and distances between centroids of estimated locations and ground truth (higher values of IoU correspond to better fit). }
\label{Table:Evaluation}
\end{table}

Let $S_i$ be the OSM perimeter of a building (polygon), and $T_{x,y}S_i$ be its translation by $x$ meters horizontally (east-west) and $y$ meters vertically (north-south).
Let $\hat{T}$ be the optimal translated position of the OSM building delivering maximum to our fit criterion IoU, where we consider overlap between the OSM-translated polygon and the 2d point cloud. Then the perimeter of the building in our point cloud can be found as $\hat {T}S_i$, with
$$\hat {T}   = \operatorname{arg\,max}_T \text{\bf \, IoU}(T_{x,y} S_i, D_{2d}), \text{\; with \;} x^2+y^2 < r^2.$$
To establish the overlap we also take into account the density of the 2d point cloud, which is non-uniform, due to substantially higher density of points along the edges of buildings (originating from points on the vertical surfaces).

\begin{figure}[t!]
\begin{center}
\includegraphics[width=.9\textwidth, height = 9cm]{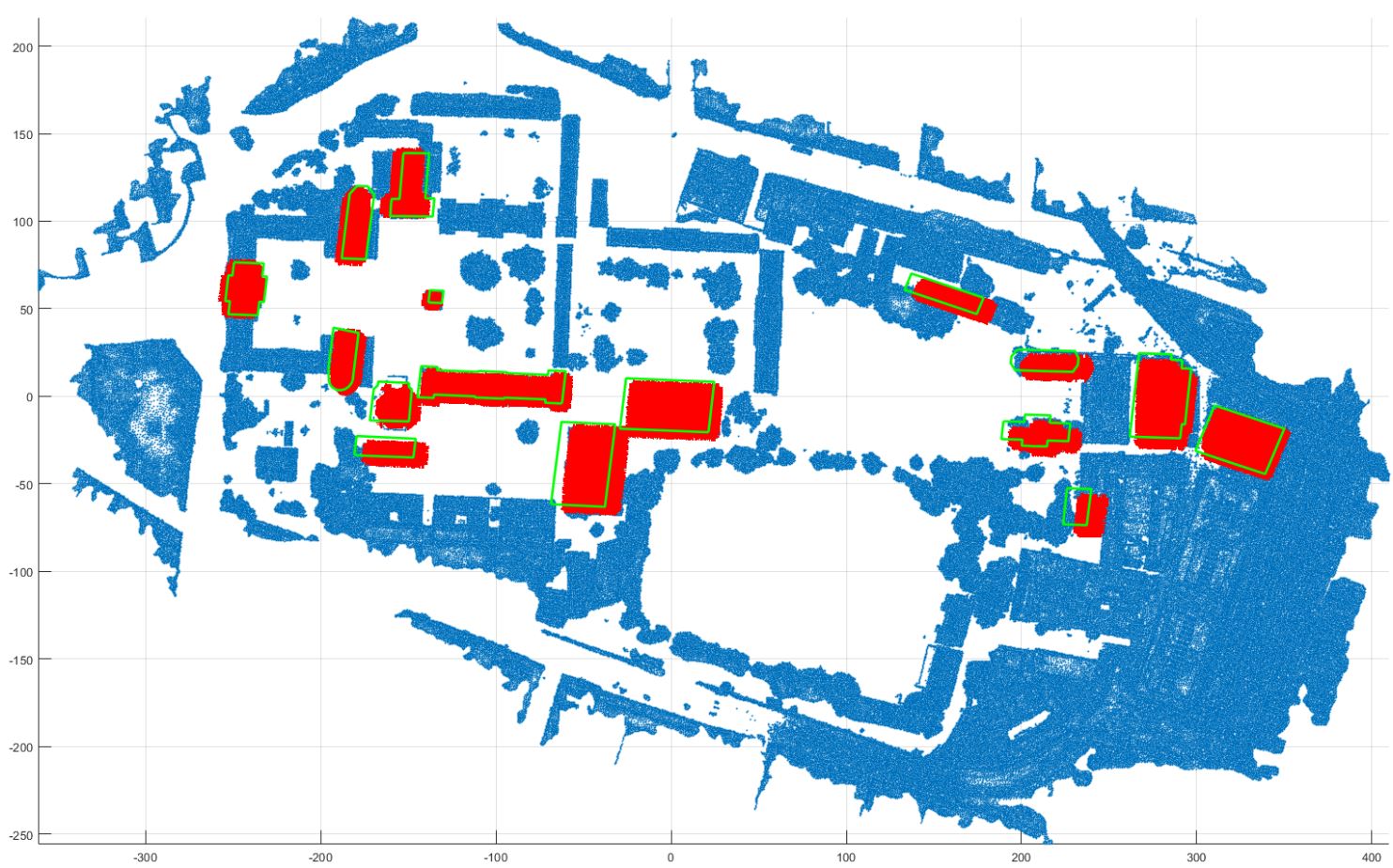}\caption{Fitting OSM to point cloud: blue patches are non-ground points from the point cloud (buildings and trees), green bounding boxes are original OSM structures, and in red are the estimated adjusted OSM positions.}
\label{fig:result}
\end{center}
\end{figure}

Practically we simplify the above maximization problem by considering fixed sizes of increments for $x$ and $y$ translations in $T_{x,y}$.
We apply two step sizes $\Delta_1=1.15m$ and $\Delta_2=0.8m$, and conduct the search of the optimal position within the $r_1=12m$, $r_1=8m$ and $r_3=5.5m$. The overall result for $(r_1,\Delta_1)$ is demonstrated in Fig.~\ref{fig:result}. The green polygon represents the initial OSM polygon, and the red data points are the segmented parts of the point cloud identified by our procedure.

In Fig.~\ref{buildings} we demonstrate the performance of adjustment procedure with the maximal radius increasing from $r_3=5.5m$ to $r_1=12m$ with a spatial translation step set to $\Delta_1=1.15m$. The initial OSM position for both highlighted buildings has a limited overlap with the point cloud data.
For the Zoology building the IoU increases significantly from $0.15$ (OSM after global registration) to $0.798$ (after adjustment in $r_3=12m$). Fig.~\ref{fig:Process_Zoology_a}-\ref{fig:Process_Zoology_c} demonstrates the gradual improvement of the OSM position with the increase of radius. In case of reasonably isolated buildings with little vegetation around this behaviour is typical. FitzGerald buildings in Fig.~\ref{fig:Process_FitzGerald_a}-\ref{fig:Process_FitzGerald_c} is an example of an isolated building closely surrounded by trees, which limits the performance of the position refinement procedure. Specifically, after the initial improvement from $r_3=5.5m$ to $r_2=8m$, the estimated position in $r_1=12m$ drops due to an adjacent cluster of dense tall vegetation which biases the estimated position.

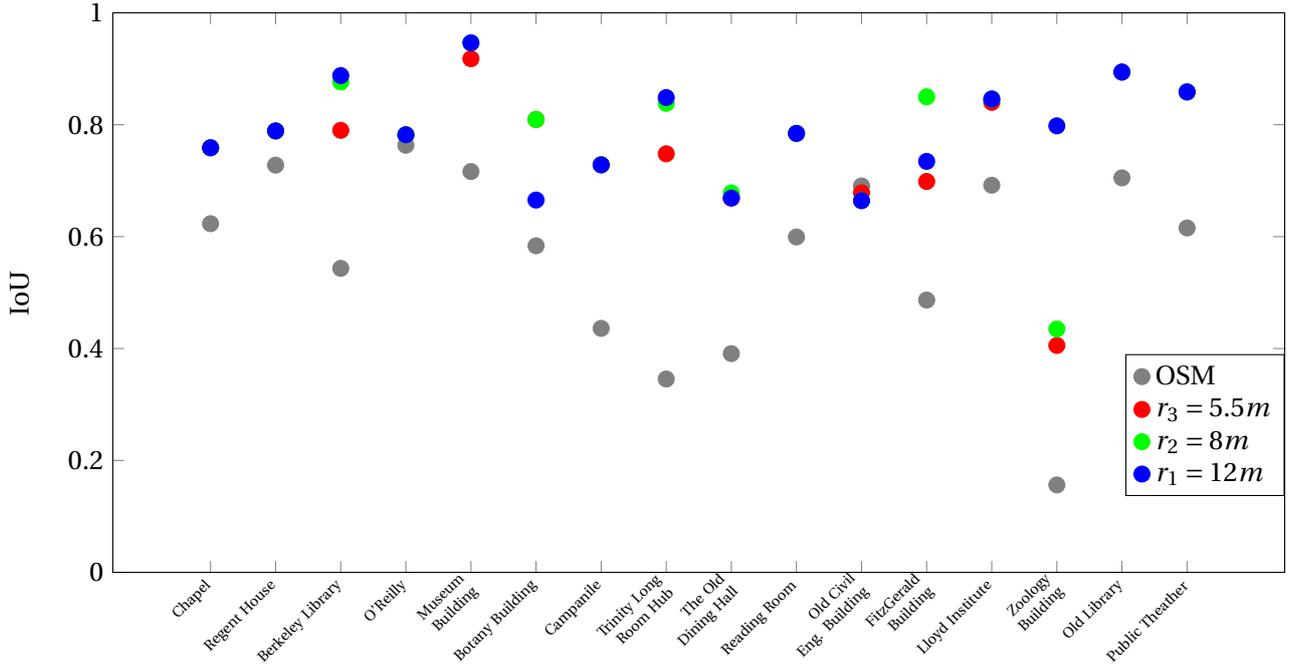
\begin{figure}[!t]
\begin{tikzpicture}
    \begin{axis}
        [
        ymin=0, ymax=1,
        ,width=\textwidth
        ,height = 9cm
        ,ylabel=IoU
        ,xtick=data,
         xticklabel style = {rotate=45, anchor=east, align=right,text width=1.5cm,font=\tiny,yshift=-.5ex},
        ,xticklabels={Chapel, Regent House, Berkeley Library, O'Reilly, Museum Building, Botany Building, Campanile, Trinity Long Room Hub, The Old Dining Hall, Reading Room, Old Civil Eng. Building, FitzGerald Building, Lloyd Institute, Zoology Building, Old Library, Public Theather},
        legend style={at={(0.932,0.39)},anchor=north,legend cell align={left}}
        ]
      \addplot[
        only marks,
        mark =*,
        color= gray,
        mark options={fill=gray},
        mark size=3pt]
    	coordinates
		{(0,0.6231)(1,0.7278)(2,0.5433)(3,0.7633)(4,0.7163)(5,0.5837)(6,0.436)(7,0.3455)(8,0.3909)(9,0.5994)(10,0.6906)(11,0.4867)(12,0.6919)(13,0.1562)(14,0.705)(15,0.6155)};   
        \addplot[
        only marks,
        mark =*,
        color= red,
        mark options={fill=red},
        mark size=3pt]
    	coordinates
		{(0,0.7588)(1,0.7889)(2,0.79)(3,0.782)(4,0.9179)(5,0.8093)(6,0.7283)(7,0.748)(8,0.669)(9,0.7845)(10,0.6782)(11,0.6986)(12,0.84)(13,0.4055)(14,0.8941)(15,0.8586)};
        
        \addplot[
        only marks,
        mark =*,
        color= green,
        mark options={fill=green},
        mark size=3pt]
    	coordinates
		{(0,0.7588)(1,0.7889)(2,0.8763)(3,0.782)(4,0.9463)(5,0.8093)(6,0.7283)(7,0.8382)(8,0.6782)(9,0.7845)(10,0.6642)(11,0.8498)(12,0.846)(13,0.4349)(14,0.8941)(15,0.8586)};
        \addplot[
        only marks,
        mark =*,
        color= blue,
        mark options={fill=blue},
        mark size=3pt]
    	coordinates
		{(0,0.7588)(1,0.7889)(2,0.8878)(3,0.782)(4,0.9463)(5,0.6654)(6,0.7283)(7,0.8485)(8,0.669)(9,0.7845)(10,0.6642)(11,0.7345)(12,0.846)(13,0.798)(14,0.8941)(15,0.8586)};
     \legend{OSM,$r_3=5.5m$,$r_2=8m$,$r_1=12m$}
    \end{axis}
\end{tikzpicture}
\caption{IoU per building: initial OSM (after global affine registration) and adjusted position with three radii $r$.}\label{plot:secondIoU}
\end{figure}

Table~\ref{Table:Evaluation} demonstrates performance of the proposed position adjustment algorithm for various values of search radius $r$ and translation step $\Delta$. We show the average and median for these 16 buildings and we found that the IoU increases significantly. From 0.567 on average, after the global registration of OSM data, to above $0.75$ after position adjustment.
%, see individual improvements in Fig.~\ref{plot:secondIoU}. 
The distance between centroid point of polygons from 5.594 meters down to around 2-2.4 meters after automatically estimated translation adjustment $\hat{T}$. In addition, in Fig.~\ref{plot:secondIoU} we demonstrate the improvement in IoU  achieved with the proposed position adjustment approach for the considered individual buildings with the translation step $\Delta_2=1.15m$ and three radii $r$. The proposed greedy search strategy consistently improves the overlap between OSM shapes and point cloud data. We implement this algorithm on CPU(Intel i7 with 8 threads) in MATLAB. The whole segmentation process takes around 6-8 minutes depending on maximum  distance $r_i$ and step $\Delta_i$.

\section{Conclusion \& Future work}

Our overall results show that aligning GIS (OSM) data with 3D point cloud is promising for segmenting it in meaningful subsets of vertices. 
We are currently working on improving the robustness of the pipeline by integrating the colour information into the decision process. 
%The pipeline can be used in different 3D real world mesh.

\section*{Acknowledgement}
This work was supported by the ADAPT Centre for Digital Content Technology, funded by the
Science Foundation Ireland Research Centres Programme (Grant 13/RC/2106) and the European Regional Development Fund.

\bibliographystyle{apalike}
\bibliography{smileBib,RozBib}

\end{document}